# High-Throughput Computational Screening and Interpretable Machine Learning of Metal-organic Frameworks for Iodine Capture in Humid Environments

Haoyi Tan[1], Yukun Teng[1] and Guangcun Shan[1,2,*]

[1]School of Instrumentation Science and Opto-electronics Engineering, Beihang University, Beijing 100083, China

[2]Department of Materials Science and Engineering, City University of Hong Kong, Hong Kong SAR, China

*e-mail: gshan2-c@my.cityu.edu.hk

## Abstract

The removal of leaked radioactive iodine isotopes in humid environments holds significant importance in nuclear waste management and nuclear accident mitigation. In this study, high-throughput computational screening and machine learning were combined to reveal the iodine capture performance of 1816 metal-organic framework (MOF) materials under humid air conditions. Firstly, the relationship between the structural characteristics of MOF materials (including density, surface area and pore features) and their adsorption properties was explored, with the aim of identifying the optimal structural parameters for iodine capture. Subsequently, two machine learning regression algorithms - Random Forest and CatBoost, were employed to predict the iodine adsorption capabilities of MOF materials. In addition to 6 structural features, 25 molecular features (encompassing the types of metal and ligand atoms as well as bonding modes) and 8 chemical features (including heat of adsorption and Henry's coefficient) were incorporated to enhance the prediction accuracy of the machine learning algorithms. Feature importance was assessed to determine the relative influence of various features on iodine adsorption performance, in which the Henry's coefficient and heat of adsorption to iodine were found the two most crucial chemical factors. Furthermore, four types of molecular fingerprints were introduced for providing comprehensive and detailed structural information of MOF materials. The top 20 most significant MACCS molecular fingerprints were picked out, revealing that the presence of six-membered ring structures and nitrogen atoms in the MOF framework were the key structural factors that enhanced iodine adsorption, followed by the existence of oxygen atoms. This work combined high-throughput computation, machine learning, and molecular fingerprints to comprehensively and systematically elucidate the multifaceted factors influencing the iodine adsorption performance of MOFs in humid environments, offering profound insightful guidelines for screening and structural design of advanced MOF materials.

## Introduction

As an efficient and low-carbon energy source, nuclear energy plays a significant role in the global energy landscape, particularly in the context of global climate change, where it provides significant support for achieving carbon neutrality goals[1, 2]. However, the rapid expansion of nuclear energy industry is accompanied by potential environmental and safety risks[3], especially in the handling of nuclear waste and during nuclear accidents, where the leakage of radioactive substances poses a severe threat to both the environment and human health. Among the radioactive isotopes involved in spent nuclear fuel reprocessing or nuclear accidents, iodine isotopes, particularly $^{131}$I and $^{129}$I, are of particular concern due to their volatility and strong bioaccumulation, which result in significant long-term impacts on the environment and human health[4]. The half-life of $^{131}$I is only 8 days; although its radiation is intense, the associated risks are generally short-term, primarily entering the human body via inhalation or the food chain, leading to acute health issues such as thyroid cancer. In contrast, $^{129}$I has an exceptionally long half-life (~$1.57 \times 10^7$ years), enabling it to persist in the biosphere and cause sustained threats to ecosystems. Consequently, the efficient removal of radioactive iodine isotopes has become an urgent requirement for ensuring nuclear safety and reducing environmental contamination[5, 6, 7, 8].

Metal-organic frameworks (MOFs), as novel porous materials formed by metal clusters coordinated with organic ligands, have gained considerable attention as potential iodine adsorbents due to their highly tunable structures, large surface areas, and excellent porosity[9, 10]. However, in the real nuclear industry and spent nuclear fuel reprocessing, high-humidity air environments are prevalent, thus demanding more robust iodine adsorption properties from MOF materials[6, 11, 12]. In recent years, many researchers have investigated the iodine adsorption behavior of various MOFs in humid environments: Tina M. Nenoff and co-workers explored the competitive $I_2$ sorption by Cu-BTC from humid gas streams (about 3.5% relative humidity) at 75°C and ambient pressure, revealing a remarkable iodine capacity of ~175 wt % with a derived $I_2/H_2O$ adsorption selectivity of 1.5[13]. Praveen K. Thallapally's group reported $I_2$ adsorption capacities and mechanisms in two microporous MOFs in the presence of humidity (33% RH and 43% RH), in which SBMOF-1 and SBMOF-2 exhibited the 15 wt % and 35 wt % uptake, respectively[14]. Zhang et al. systematically studied the influence of $H_2O$ molecules on the iodine adsorption properties of different zeolitic imidazolate frameworks (ZIFs) using grand canonical Monte Carlo (GCMC) simulations, highlighting the competitive adsorption behavior between $H_2O$ and $I_2$, particularly for hydrophilic materials[15]. Other MOF materials including MIL-101-Cr-TED, MIL-101-Cr-HMTA and ECUT-300, also were used to explore the iodine capture performance in a water-containing system[16, 17, 18, 19]. In our previous work, grand canonical Monte Carlo (GCMC) and density functional theory methods were employed to investigate the iodine adsorption performance of 21 chemically stable MOF materials in high-humidity environments, and influence of different structural factors were revealed[20]. However, despite these advances, researches on the iodine adsorption behavior of MOFs under humid conditions remains limited, and a comprehensive insight of the key factors influencing iodine adsorption based on a larger number of MOF materials is still needed.

Nowadys, high-throughput computational screening based on molecular simulations offers a rapid approach to evaluating the iodine adsorption performance of MOFs under humid conditions[21, 22]. Furthermore, the rapid development of artificial intelligence has ushered in a new research paradigm that combines data science with chemistry[23, 24]. Machine learning has proven to be an efficient tool for analyzing computational data related to gas adsorption behaviors of MOFs, for revealing structure-property relationships, identifying promising MOF adsorbents and even guiding MOF structures design and modification[25, 26, 27, 28]. In this work, we first selected 1816 $I_2$-accessible MOF materials (with pore limiting diameter > 3.34 Å - the kinetic diameter of $I_2$) from the well-established CoRE MOF 2014 database established by Chuang et al.[29], and employed GCMC simulations to study their $I_2$ adsorption performance under humid air conditions[30]. Subsequently, three different types of descriptors (structural, molecular, and chemical features) were explored, and machine learning algorithms were utilized to predict iodine adsorption performance and reveal the relationships between various descriptors and iodine adsorption. Finally, molecular fingerprint technique was employed to comprehensively identify the influence of structural features on iodine adsorption, providing valuable insights for the future molecular design of MOF materials.

## Results

### Structure-performance relationships

To identify the optimal structural features, the relationships between the 1816 MOF structures and iodine adsorption performance were investigated in Fig. 1 and Fig. S1. Structural characteristics of MOFs included the pore limiting diameter (PLD), largest cavity diameter (LCD), void fraction (φ), pore volume, surface area and density. When LCD was less than 4 Å (Fig. 1a), the spatial steric hindrance between the $I_2$ molecules and the pore walls resulted in negligible iodine adsorption. When 4 Å < LCD < 5.5 Å, an increase in LCD reduced the steric hindrance, thereby adsorption interaction between the framework materials and iodine molecules became the dominant factor, which led to an increase in both iodine adsorption capacity and selectivity. However, when LCD exceeded 5.5 Å, further enlargement of the channel size diminished the interaction between MOFs and iodine molecules, which intensified the desorption of $I_2$ in the pores and resulted in a continuous decline in both adsorption capacity and selectivity. To identify MOF materials with optimal iodine adsorption properties, we could find the ideal value for LCD lay between 4 and 7.8 Å. As for porosity (with the optimal value for iodine adsorption less than 0.17) in Fig. 1b, iodine adsorption capacity and selectivity initially increased (φ < 0.1) and then decreased (0.1 < φ < 0.6). The relationships between density and iodine adsorption performance also followed a similar trend (Fig. 1c): at low densities (under 0.9 g/cm³), the increase in density promoted iodine adsorption due to the greater number of available adsorption sites; however, when the density exceeded 0.9 g/cm³, excessively compact pore structures limited iodine adsorption, with the attraction interaction between the MOFs and $I_2$ molecules changing to repulsion interaction; with the density further surpassing 2.2 g/cm³, iodine uptake amount would fall below 100 cm³/g. Furthermore, the optimal values of pore volume, PLD and surface area for iodine capture were also identified, which lay at the range of 0 ~ 0.18 cm³/g, 3.34 ~ 7 Å and 0 ~ 540 m²/g, respectively (Fig. 1d and Fig. S1). In order to facilitate the comparison of the adsorption behavior of different molecules, structure-performance relationships of MOF materials for $H_2O$ adsorption have been delineated (Fig. S2), in which the optimal structural parameters exhibited a broader range (for instance, the optimal LCD could reach up to 11 Å and the optimal φ could attain 0.48); this is likely attributed to the larger kinetic diameter of $H_2O$ molecules compared to $I_2$. The above results explained that relatively small pore sizes of MOF materials could confer the advantage during competitive iodine adsorption in humid conditions.

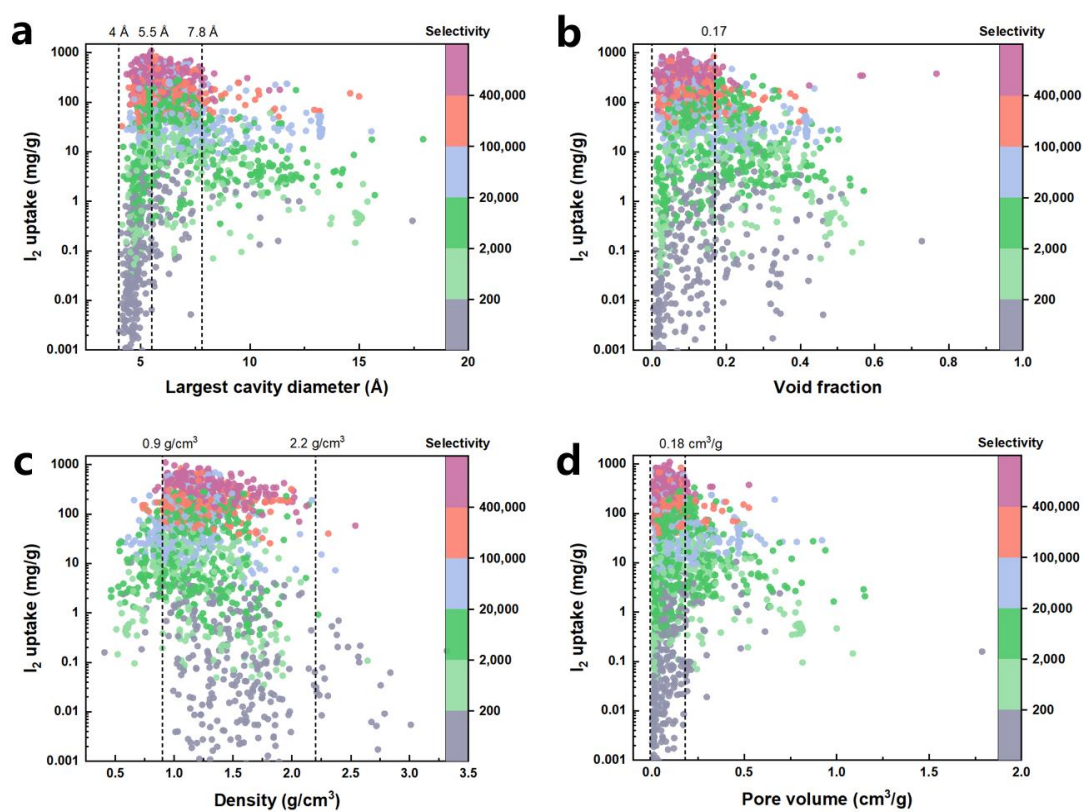

**Fig. 1 | Structure-performance relationship.** $I_2$ capture performance (uptake amount and selectivity) as a function of **a** largest cavity diameter (Å), **b** void fraction, **c** density (g/cm³) and **d** pore volume (cm³/g).

## Machine learning

After the aforementioned analysis, we initially employed six structural descriptors - PLD, LCD, φ, pore volume, surface area and density, to train machine learning algorithms for predicting iodine gas adsorption in MOF materials under humid conditions (Fig. 2a, d). Two different machine learning algorithms (including random forest and CatBoost model) were trained and compared[31,][32]. After training the model with only structural parameters as the simplest feature set, we gradually incorporated more comprehensive feature sets, including "structural + molecular descriptors" (Fig. 2b, e) and "structural + molecular + chemical descriptors" (Fig. 2c, f). For molecular descriptors, each molecular feature corresponded to specific elemental, hybridization, and bonding types. For carbon (C) and nitrogen (N) elements, the atomic types included C_1, C_2, C_3, and C_R (or N_1, N_2, N_3, and N_R), depending on the nature of single, double, triple, and ring bonds. Oxygen (O) atoms can form double bonds and ring bonds, defined as O_2 and O_R, respectively, as well as central tetrahedral oxygen (denoted as O_3_f) or central trigonal oxygen (denoted as O_2_z)[33]. For hydrogen (H), fluorine (F), chlorine (Cl), and bromine (Br), the atomic types are designated as H_, F_, Cl_, and Br_. Additionally, tetrahedral four-coordinate phosphorus for organo-metallic coordination is defined as P_3+q, along with sulfur atoms connected via cyclic bonds (denoted as S_R)[34]. Regarding metal atoms, only the predominant metal species within the MOF were considered: descriptors for these metals included metal ratio (the molar ratio relative to all atoms), atomic number, atomic weight, atomic radius, polarizability, electron affinity, and Mulliken electronegativity. In chemical descriptors, Henry's coefficient and heat of adsorption of $I_2$, $H_2O$, $N_2$ and $O_2$ in MOF materials were considered and defined as $I_2$_Henry ($I_2$_heat), $H_2O$_Henry ($H_2O$_heat), $N_2$_Henry ($N_2$_heat) and $O_2$_Henry ($O_2$_heat), respectively.

Throughout the above process, the prediction performance of both the random forest and CatBoost algorithms progressively improved. The accuracy of the machine learning was evaluated using $R^2$, mean absolute error (MAE), and root mean square error (RMSE). When only structural descriptors were used, the random forest algorithm exhibited a relatively low prediction accuracy ($R^2$ = 0.438). After adding molecular descriptors to the feature set, the prediction accuracy of the random forest model improved, with $R^2$ increasing to 0.592. Further incorporation of chemical descriptors led to the highest prediction accuracy ($R^2$ = 0.900). Simultaneously, the MAE and MSE values of random forest model demonstrated a steady decrease: MAE reduced from 75.588 to 61.673, and finally to 23.378; MSE decreased from 14,293.744 to 10387.059, and ultimately to 2547.433. For the CatBoost algorithm, the trends were similar to those observed with the random forest model, but overall, it exhibited better prediction performance. When using the "structural descriptors + molecular descriptors + chemical descriptors" feature set, the CatBoost algorithm achieved the highest $R^2$ of 0.941, with MAE and MSE dropping to 18.276 and 1512.681, respectively. Additionally, by following the above process, the accuracy of the CatBoost algorithm in predicting the $H_2O$ adsorption performance of MOF materials also gradually improved (Fig. S3), and the value of $R^2$, MAE and MSE respectively reached 0.911, 5.672 and 166.554 based on "structural descriptors + molecular descriptors + chemical descriptors" feature set. This further confirmed that molecular and chemical descriptors significantly complemented the structural descriptors, playing a crucial role in accurately predicting gas molecules capture in MOF materials.

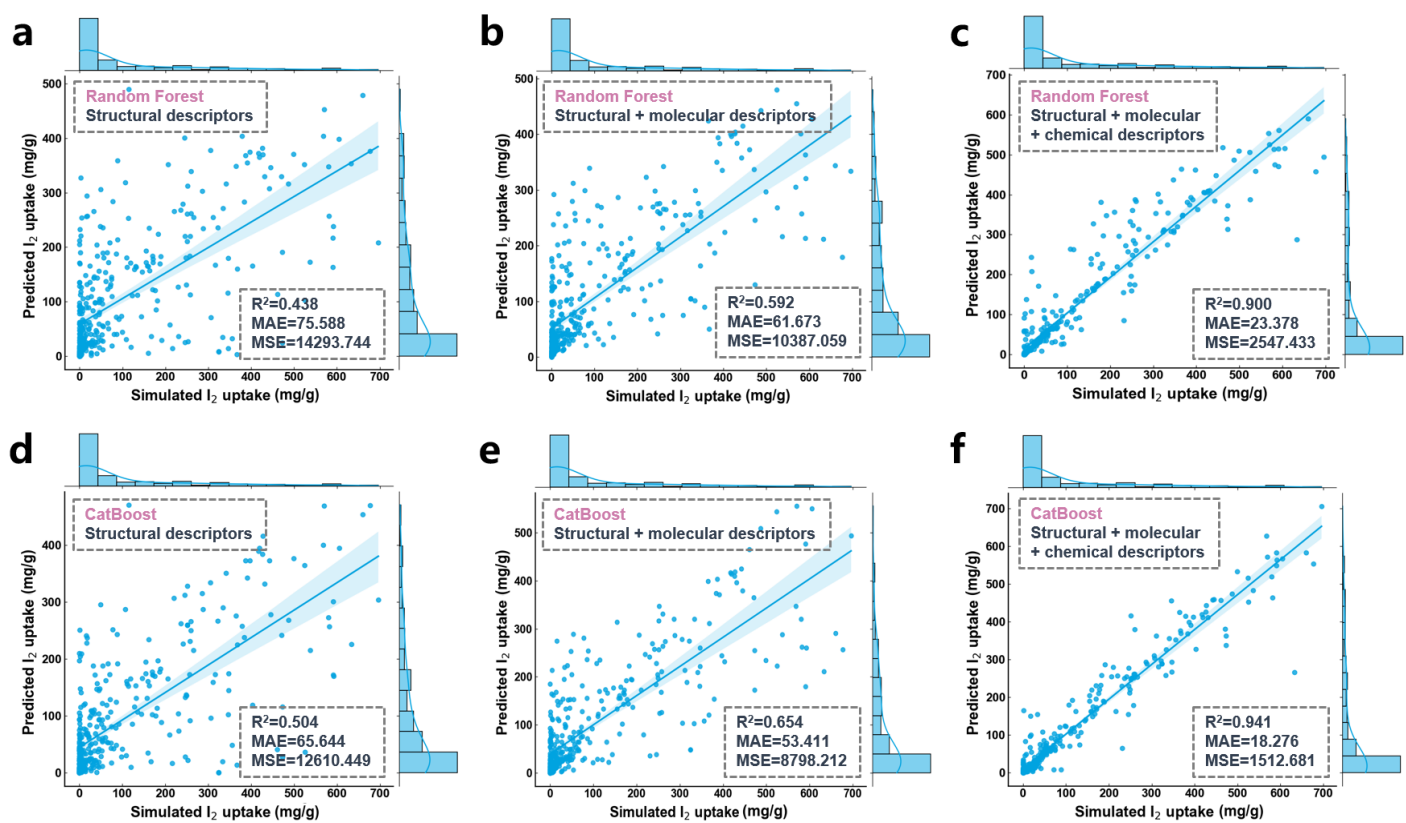

**Fig. 2 | Comparisons of prediction performance.** Prediction accuracy of I$_2$ capture using random forest and CatBoost algorithms based on **a**, **d** "structural descriptors", **b**, **e** "structural + molecular descriptors" and **c**, **f** "structural + molecular + chemical descriptors" feature sets.

To investigate the contribution of different features in predicting iodine adsorption performance, the SHAP (SHapley Additive exPlanations) method was used to rank and explain the significance of various features in the CatBoost model (Fig. 3). Among the structural descriptors (Fig. 3a), LCD was identified as the most important descriptor, followed by void fraction and surface area (both of which exhibit a certain negative correlation with I$_2$ adsorption performance). PLD and density of the MOF material ranked next in importance, while pore volume had the least significance among the six structural descriptors. In the "structural descriptors + molecular descriptors" set (considering only the top 20 descriptors in terms of importance), all structural descriptors, except for pore volume, ranked within the top 10, with LCD maintaining the highest importance. Among the molecular descriptors, the most significant was C_R (positively correlated with adsorption capacity), followed by the proportion of metal atoms (negatively correlated with adsorption capacity), and then H_R, N_R, and O_2 (all positively correlated with adsorption capacity). We speculated that iodine adsorption in MOF materials primarily relied on the organic ligands, with carbon rings, nitrogen, and oxygen atoms serving as key adsorption sites. In the "structural descriptors + molecular descriptors + chemical descriptors" set (considering only the top 20 descriptors), the most influential features were I$_2$_Henry, followed by I$_2$_Heat. Both features were positively correlated to iodine capture amounts. Notably, in contrast to N$_2$ and O$_2$, the adsorption heat of H$_2$O (H$_2$O_heat) was strongly negatively correlated with iodine adsorption capacity (with the damping coefficient positively correlated), likely because under humid conditions, H$_2$O molecules are the main competitive species for iodine gas adsorption. In terms of H$_2$O molecules adsorption (Fig. S4), H$_2$O_heat possessed the highest relative importance, and metal atoms also exhibited relatively high significance with polarizability, metal ratio, Mulliken electronegativity and atomic radius positioning within the top six rankings. In contrast to the adsorption of I$_2$ molecules, the atomic radius of the metal atoms exhibited a positive correlation with the adsorption of H$_2$O molecules (Fig. S4b); furthermore, O_R replaced N_R as the most favorable ligand structure due to the strong hydrogen bonding interactions (Fig. S4c).

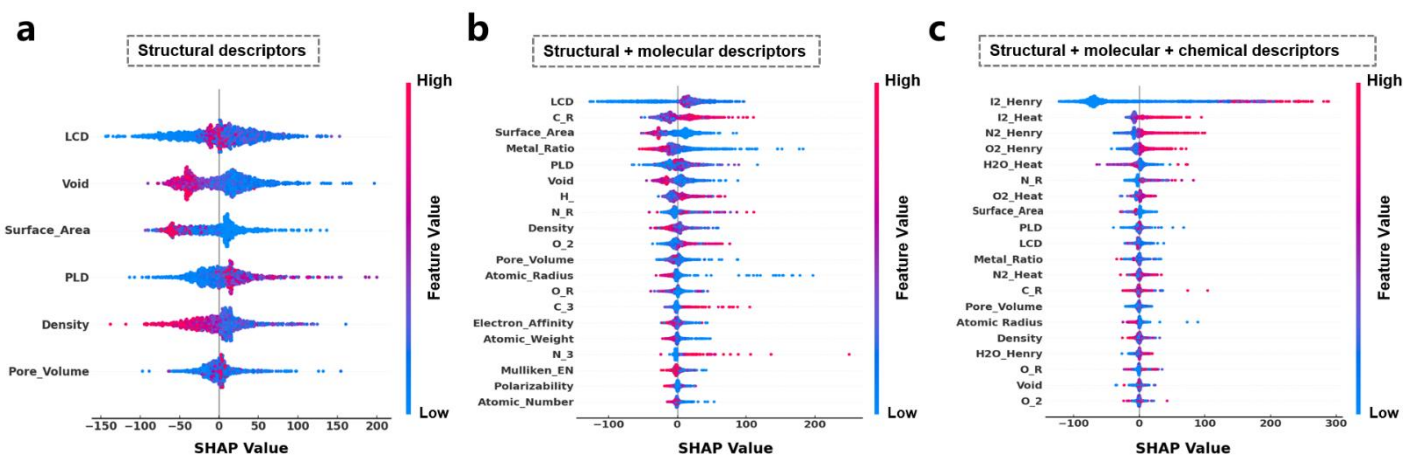

**Fig. 3 | Feature importance for I$_2$ capture.** SHAP value distribution of **a** "structural descriptors", **b** "structural + molecular descriptors" and **c** "structural + molecular + chemical descriptors".

The correlation coefficients between different features were calculated and visualized as heatmaps. Among the pure structural descriptors (Fig. S5), a strong positive correlation (correlation coefficient > 0.6) was evident between PLD, LCD, void fraction, surface area, and pore volume, while density showed a strong negative correlation (correlation coefficient > -0.4) with the other five structural parameters. In the "structural descriptors + molecular descriptors" set (Fig. S6), metal ratio, metal atomic number, metal atomic weight, and metal atomic radius exhibited positive correlations with the density of the MOF material (with correlation coefficients of 0.32, 0.38, 0.39, and 0.28, respectively), due to the heavier nature of metal atoms compared to ligand atoms. Additionally, the metal atoms ratio was positively correlated with void fraction (with correlation coefficient of 0.23), whereas the correlations between porosity and the number of metal atoms, metal atomic weight, and metal atomic radius were relatively weak. This was because metal clusters, formed by a greater number of metal atoms as connecting nodes, would lead to the increased porosity, irrespective of the type of metal atoms. The metal ratio was negatively correlated with the number of most organic ligand atoms (including H_R, C_R, N_R, O_R, C_3, and N_3) (with correlation coefficient of -0.43, -0.42, -0.11, -0.10, -0.15, and -0.13), but was positively correlated with O_2 (with correlation coefficient of 0.31), which was likely due to the fact that a certain amount of O existed to form metal clusters in MOF materials by bonding with metals atoms. In the "structural + molecular + chemical descriptors" set (Fig. 4), $I_2$_Henry showed a low correlation with both structural and molecular descriptors. $N_2$_Henry, $O_2$_Henry, and $H_2O$_Henry exhibited significant positive correlations with most structural descriptors (including PLD, LCD, porosity, specific surface area, and pore volume), while $N_2$_Heat, $O_2$_Heat, and $H_2O$_Heat were negatively correlated with the aforementioned descriptors. This was attributed to the fact that the larger size of cavity or pore channel in MOFs facilitated the diffusion of gas molecules within MOFs, but the reduced density of adsorption sites weakened the adsorption strength, leading to the lower adsorption heats. The very strong correlations between $N_2$_Henry and $O_2$_Henry (correlation coefficient = 0.98) and between $N_2$_Heat and $O_2$_Heat (correlation coefficient = 0.89) arose from the similar molecular structures of $N_2$ and $O_2$. $I_2$_Heat was negatively correlated with the proportion of metal atoms (correlation coefficient of -0.27) and positively correlated with C_R (correlation coefficient of 0.21), further indicating that metal sites were not effective adsorption sites for $I_2$ molecules, which tended to be adsorbed near organic ligands (such as benzene rings).

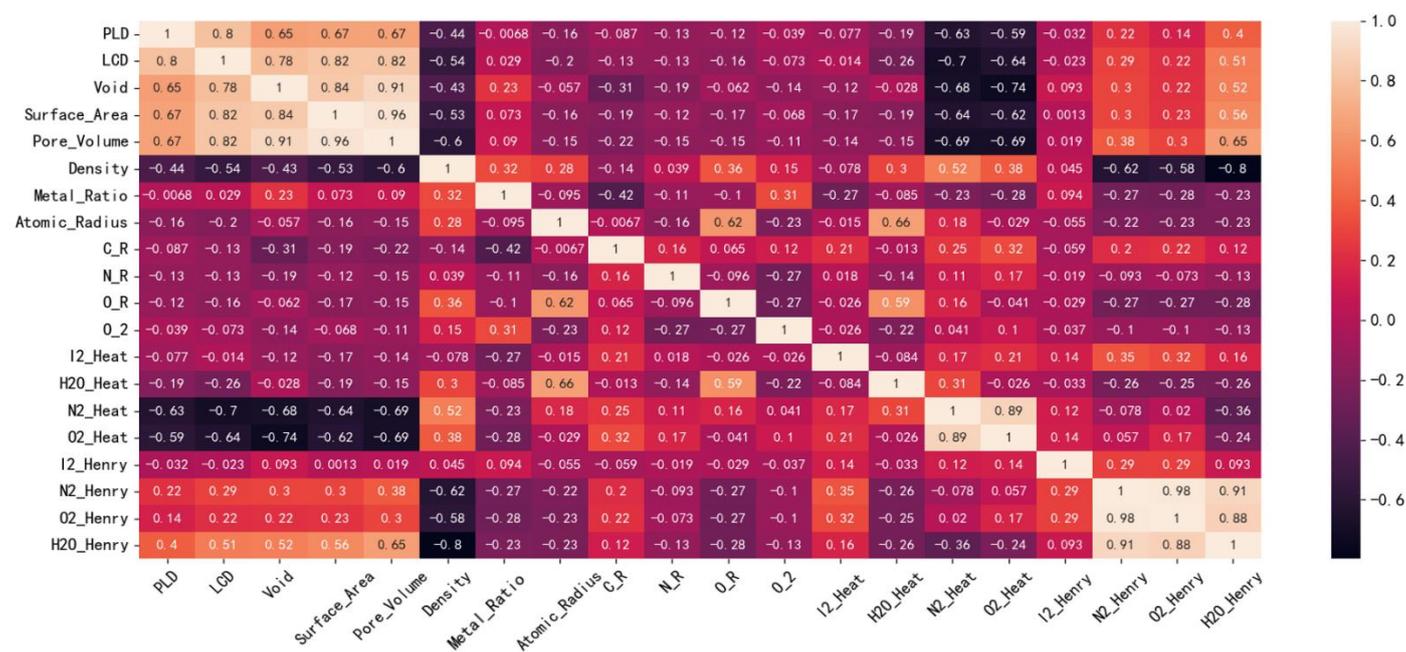

**Fig. 4 | Heatmap of correlation coefficients.** Pearson coefficients matrix for the 20 most significant features for $I_2$ capture in "structural + molecular + chemical descriptors" feature set.

## Molecular Fingerprint

In order to comprehensively reveal the factors that positively or negatively influenced iodine adsorption performance, thereby paving the way for future molecular design, four types of molecular fingerprint (including Molecular ACCess Systems (MACCS), PubChem, AtomPairs2D and Estate fingerprint) were employed in place of the previously used molecular descriptors. MACCS and PubChem represented two of the most widely used fingerprint derived from substructure key information: the MACCS fingerprint consisted of a set of 166 structural keys, constructed using SMART patterns; while the PubChem fingerprint originated from the

PubChem database, encompassing 881 types of structural keys represented as binary substructure encodings. The APFP fingerprint encoded a total of 780 atomic pairs based on their topological distances. Estate summarized the microscopic structure of materials through a 79-byte representation. Based on the presence or absence of bits, molecular fingerprint digitized the molecular features of MOF materials, thereby providing microscopic insights into the structure of exceptional materials. These molecular fingerprints, in conjunction with previous structural and chemical descriptors, were applied to train machine learning models for the prediction of iodine adsorption performance under humid conditions. Although the prediction performance showed a slight decrease based on "structure + molecular fingerprint + chemical descriptors" set compared to the prior "structure + molecular + chemical descriptors" set, because the encoding of molecular fingerprint solely indicated the presence or absence of specific features and other information such as the quantity or proportion in a single MOF unit cell were missed, which limited their ability in prediction of uptake amount; however, their comprehensive inclusion of various structural categories enabled them to serve as excellent interpretative tools.

After comparing the prediction accuracies of the four molecular fingerprints (Fig. 5a and Fig. S7), MACCS molecular fingerprint exhibited its superiority in machine learning ($R^2 = 0.927$, MAE = 20.057, and MSE = 1651.391). The 20 most significant MACCS molecular fingerprints were ranked and accompanied by detailed interpretations (Fig. 5b and Table S1). Additionally, the autocorrelation coefficients of the molecular fingerprints were presented in the heatmap to reveal the interrelationships among the fingerprints (Fig. S8). The top two, Bit_158 and Bit_75, demonstrated the significant positive impact of nitrogen (N) atoms in MOF materials on iodine adsorption performance. The strong correlations between Bit_158 and Bit_156 (with correlation coefficient of 0.9), and between Bit_158 and Bit_45 (with correlation coefficient of 0.73) further highlighted that the presence of N atoms, whether in rings or directly bonded to carbon (C) atoms, promoted the iodine adsorption in humid air conditions. Bit_163 and Bit_162, which had a correlation coefficient of 0.6, represented that six-membered aromatic ring was another important structural feature. Six-membered aromatic ring likely provided electron-rich adsorption sites due to their large π-bonds, thereby enhancing the iodine adsorption. Bit_6 and Bit_12, which were associated with the type of metal, represented lanthanide metal and group IB (or IIB) metal elements, respectively. The negative correlation (-0.38) between these two molecular fingerprints arose from the relatively homogeneous nature of the metal clusters in these materials; however, the former demonstrated a negative effect on $I_2$ adsorption, while the latter had a positive impact (consistent with previous findings, where a negative correlation was observed between the metal atomic radius and iodine adsorption performance). The relationships between MACCS molecular fingerprint and other structural (or chemical) features were also illustrated (Fig. S8). It could be found that although Bit_6 (lanthanide metals) and Bit_12 (group IB/IIB metals) showed little correlation with $I_2$_Henry and $I_2$_Heat, Bit_6 enhanced the adsorption of $H_2O$ in MOF materials (correlation coefficient with H2O_Heat of 0.65), while Bit_12 weakened the adsorption of $H_2O$ (correlation coefficient with H2O_Heat of -0.41); thus, it could be concluded that group IB/IIB metals played the significant role in promoting competitive adsorption of $I_2$ under humid conditions compared to lanthanide metals.

Furthermore, the MACCS fingerprints Bit_138, Bit_69, Bit_100 and Bit_131, representing different forms of hydrogen (H) (whether connected to C or non-C elements), and Bit_143 and Bit_139, representing different forms of oxygen (O) (either as part of a ring structure or as hydroxyl groups), all demonstrated the positive role of H and O atoms in enhancing iodine adsorption in MOF structures. Nevertheless, modifications involving N atoms appeared to offer the greater advantages over O atoms (Overall, Bit_158 and Bit_75 had the higher importance ranking compared to Bit_143 and Bit_139). This was because the existence of N atoms, compared to O atoms, exerted the evident suppressive effect on both pore size and the adsorption of $H_2O$ molecules, both of which were more favorable for iodine adsorption in humid environments; specifically, the correlation coefficients between N-related molecular fingerprints (Bit_158, Bit_75, Bit_156, and Bit_45) and pore volume were -0.27, -0.28, -0.26, and -0.26, respectively, and their correlations with H2O_Heat were -0.12, -0.18, -0.13, and -0.17; in contrast, O-related molecular fingerprints (Bit_143 and Bit_139) had the positive correlation coefficients with pore volume of 0.2 and 0.046, respectively, and correlation coefficients with H2O_Heat were 0.24 and 0.029. The role of H atoms lay between that of the N and O atoms: the correlation coefficients between H-related molecular fingerprints (Bit_138, Bit_69, Bit_100 and Bit_131) and pore volume were at the range of -0.1 ~ -0.2, and the correlation with H$_2$O_Heat were also weak (with negative correlation coefficients less than -0.1).

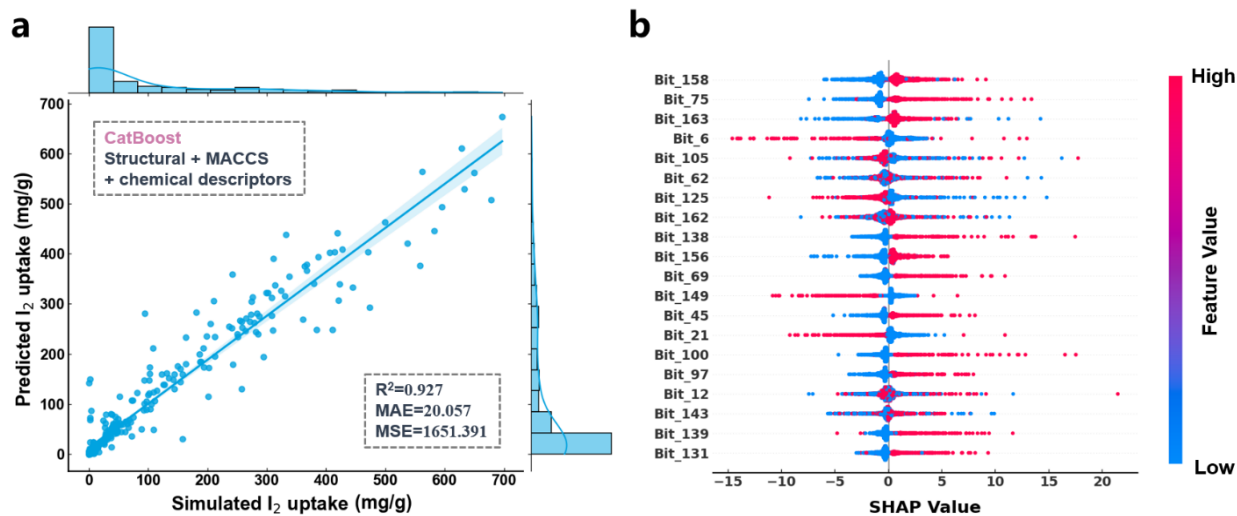

**Fig. 5 | Prediction performance and feature importance using CatBoost algorithm. a** Prediction accuracy of I$_2$ capture and **b** SHAP value distribution based on "structural + MACCS + chemical descriptors" set.

Finally, the top six MOF materials with the best iodine adsorption performance were picked out and microscopic insights of public fingerprints were provided (Fig. 6 and Table S2). These MOF structures shared a common set of molecular fingerprints: Bit_45, Bit_75, Bit_156, Bit_158, and Bit_163, all of which indicated the existence of six-membered rings and N atoms, and N atoms were part of the six-membered rings or directly coordinated with metal atoms. Furthermore, the metal atoms in these structures were all transition metals from the fourth period (including Zn, Co, Ni, and Mn), which were characterized by the smaller atomic radii and atomic numbers compared to lanthanide elements. Four of these MOF materials had Bit_97 molecular fingerprint, representing the presence of O atoms. The above findings validated the previous analysis and could aid in the future design of high-performance MOF materials.

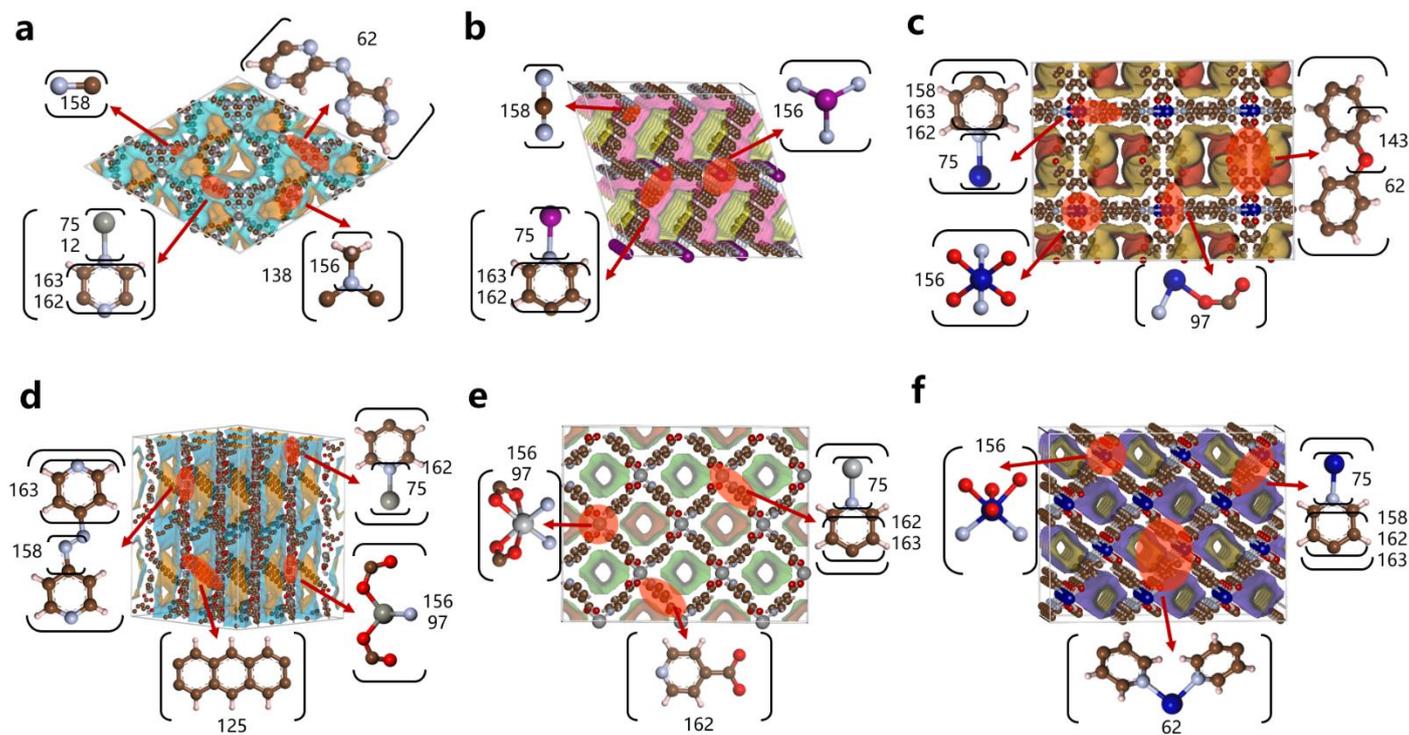

**Fig. 6 | Decomposition diagrams of molecular fingerprint of the top six MOF materials with the best iodine adsorption**

**performance.** Molecular fingerprint of MOF materials of **a** BARZUR, **b** CUVGOQ, **c** ZEXKUK, **d** QUDJOP, **e** UFATEA01 and **f** CEYPUT.

# Conclusions

In summary, large-scale GCMC simulations were employed to investigate iodine adsorption performance (including adsorption capacity and selectivity) for 1816 MOF materials from CoRE MOF database under humid conditions. Two machine learning algorithms, Random Forest and CatBoost, were utilized to predict the iodine adsorption performance of MOF materials, gradually incorporating three different types of descriptors to enhance the prediction accuracy of the models: 6 structural features (including pore characteristics, density, and surface area), 25 molecular features (including the types of metal and ligand atoms as well as their bonding modes), and 8 chemical features (including adsorption heat and Henry's coefficient). SHAP method was used to rank the importance of these descriptors, and correlation coefficients were employed to reveal the relationships among features. Four types of molecular fingerprints were also generated in place of molecular features and combined with the CatBoost algorithm to predict iodine adsorption performance. The top 20 MACCS molecular fingerprints were extracted, demonstrating the most significant role of six-membered rings and N atoms in MOF materials, followed by O atoms. Among the metal sites, the lighter transition metal elements were found to be more favorable for iodine adsorption compared to lanthanide elements. This comprehensive and systematic study shed light on the iodine adsorption performance of MOF materials under humid conditions, providing valuable insights for the future screening and design of high-performance MOF materials.

# Methods

## Simulation method

GCMC simulations were employed using RASPA software to investigate the adsorption behavior of $I_2$ within the MOFs at an environment of 423 K and 1 bar. The temperature of 423 K was relevant to operational conditions in the nuclear industry[6, 35, 36]. To replicate the high humidity environment encountered during the post-treatment phase of spent nuclear fuel, the mixed gas system was composed of 300 ppm $I_2$, 68.5% $N_2$, 18.4% $O_2$, and 12.2% $H_2O$, achieving a relative humidity of 100%[6]. Throughout the simulation, the MOFs were treated as fixed rigid structures with periodic boundary conditions. Supercells were utilized as necessary to ensure that the system dimensions exceeded twice the cutoff distance (12 Å). In addition, the selectivity of $I_2$ during adsorption was calculated as the following equation[8, 36]:

$$selectivity_{I_2} = \frac{X_{I_2}/Y_{I_2}}{X_{others}/Y_{others}}$$

where $X_{I_2}$ and $Y_{I_2}$ denoted the uptake amounts and gas phase concentration of $I_2$; $X_{others}$ and $Y_{others}$ were the uptake amounts and gas phase concentration of other gas components ($N_2$, $O_2$ and $H_2O$).

All GCMC simulations comprised an equilibration phase of 50,000 cycles, followed by a production phase of 50,000 cycles. Each cycle involved the movement of all adsorbed molecules, encompassing the insertion, deletion, translation, rotation, reinsertion, identity change, and swap processes. Iodine molecules were modeled as spherical entities, with van der Waals parameters derived from the viscosity of pure iodine[37]. Water molecules were represented using the transferable intermolecular potential (TIP3P) model ($r_{OH}$ = 0.9527 Å and $\theta_{\angle HOH}$ = 104.52°), which was a model empirically validated to accurately describe hydrogen bonding interactions[38, 39]. $N_2$ and $O_2$ molecules were modeled as three-site representation[40, 41]. The relevant molecular model parameters were referenced from previous published work[20]. The Universal Force Field (UFF) was employed to establish the Lennard-Jones (LJ) parameters for the MOF structures[42]. Interatomic interactions were described using LJ and electrostatic potential energy functions

as below:

$$U(r_{ij}) = \sum 4\varepsilon_{ij}\left[\left(\frac{\sigma_{ij}}{r_{ij}}\right)^{12} - \left(\frac{\sigma_{ij}}{r_{ij}}\right)^{6}\right] + \sum \frac{q_i q_j}{4\pi\varepsilon_0 r_{ij}}$$

where $U(r_{ij})$ denoted the non-bonded interaction energy between atoms i and j; the first term represented the van der Waals non-bonded potential energy, and the second term accounted for the Coulombic electrostatic interaction energy. $r_{ij}$ signified the interatomic distance, $\sigma_{ij}$ was the depth of the Lennard-Jones (LJ) potential well, $q_i$ and $q_j$ represented the partial charges of atoms $i$ and $j$, respectively, and $\varepsilon_0$ was the vacuum dielectric constant.

## Material descriptors

For structural descriptors, PLD and LCD were computed using the Zeo++ software package, based on Voronoi tessellation[43]. The void fraction was calculated using the RASPA software package with helium atoms (kinetic diameter = 2.58 Å) as probes. Meanwhile, the surface area, pore volume, and density were determined by RASPA using nitrogen molecules (kinetic diameter = 3.64 Å) as probes. To identify the molecular descriptors within the MOF structure, the Python program lammps_interface was utilized based on the UFF4MOF force field[33, 44, 45]. As for the chemical descriptors, adsorption heat and Henry's coefficient were calculated under infinite dilution conditions using RASPA based on an NVT-MC system (MC referred to the Monte Carlo method).

The OpenBabel and PaDEL-Descriptor software were performed to compute four types of molecular fingerprint[46, 47]: MACCS, PubChem, AtomPairs2D and Estate[48, 49, 50, 51]. OpenBabel was an open-source chemical toolbox designed to convert CIF files into SDF files compatible with PaDEL-Descriptor. And the PaDEL-Descriptor software processed the SDF format structural files, ultimately yielding the required four types of molecular fingerprint.

## Machine learning

Two machine learning regression algorithms - Random Forest and CatBoost, were implemented in Python 3.9 using the scikit-learn package[31, 32, 52]. Both of these algorithms offered advantages in terms of low computational costs and good interpretability compared to other machine learning model such as neural network algorithm[53]. Additionally, for optimizing and validating the models, we tuned the related hyperparameters using cross-validation. The accuracy of the models was evaluated using R², MAE, and MSE, whose relevant equations were as follows:

$$R^2 = 1 - \frac{\sum_{i=1}^{n}(Y_i - \widehat{Y_i})^2}{\sum_{i=1}^{n}(Y_i - \bar{Y})^2}$$

$$MAE = \frac{1}{n}\sum_{i=1}^{n}|Y_i - \widehat{Y_i}|$$

$$MSE = \frac{1}{n}\sum_{i=1}^{n}|Y_i - \widehat{Y_i}|^2$$

where $n$ represented the number of instances in the training or testing set, $Y$ denoted the predicted values from the machine learning algorithms, $\bar{Y}$ signified the mean of the model's predictions, and $\widehat{Y}$ represented the computed values for the MOF materials.

# Acknowledgements

The work was carried out at National Supercomputer Center in Tianjin, and the calculations were performed on Tianhe new generation supercomputer. This work was financially supported by the National Key R&D Program of China (No. 2016YFC1402504).

# Author contributions

H.Y.T. designed and implemented the workflow, performed the computation, generated the data, and prepared the manuscript. Y.K.T. provided importance advice and revised the paper. G.C.S. gave scientific and technical advice throughout the work, reviewed the manuscript, and supervised the project. All authors proofread and approved the final version of the manuscript.

# Competing interests

The authors declare no competing interests.

# Additional information

**Supplementary information** The online version contains supplementary material available at https://doi.org/xxxxxx

**Correspondence** and requests for materials should be addressed to Guangcun Shan